\documentclass[preprint]{article}
\PassOptionsToPackage{table}{xcolor}
\usepackage{aaai2027}
\usepackage[hyphens]{url}
\usepackage{graphicx}
\usepackage{natbib}
\usepackage{caption}
\usepackage{subcaption}
\usepackage{makecell}
\usepackage{multirow}
\usepackage{arydshln}
\usepackage{booktabs}
\usepackage{array}
\usepackage{tabularx}
\usepackage{amsmath,amsthm,amssymb}
\usepackage{adjustbox}
\usepackage{microtype}

\theoremstyle{plain}

\theoremstyle{definition}

\theoremstyle{remark}

\title{Open-Set Supervised 3D Anomaly Detection: An Industrial Dataset and a Generalizable Framework for Unknown Defects}
\author{
        Hanzhe Liang\equalcontrib\textsuperscript{\rm 1,3},
    Luocheng Zhang\equalcontrib\textsuperscript{\rm 2},
    Junyang Xia\textsuperscript{\rm 2},
    HanLiang Zhou\textsuperscript{\rm 2},\\
    Bingyang Guo\textsuperscript{\rm 1}\corresponding,
    Yingxi Xie\textsuperscript{\rm 3},
    Can Gao\textsuperscript{\rm 4},
    Jinbao Wang\textsuperscript{\rm 5},
    Ruiyun Yu\textsuperscript{\rm 6},
    Pan Li\textsuperscript{\rm 2}
}

\affiliations{
    \textsuperscript{\rm 1}
    Institute of AI for Industries,
    Chinese Academy of Sciences, Nanjing, China\\
    \textsuperscript{\rm 2}
    School of Computer Science,
    Hangzhou Dianzi University, Hangzhou, China\\
    \textsuperscript{\rm 3}
    Financial Technology Institute,
    Shenzhen University, Shenzhen, China\\
    \textsuperscript{\rm 4}
    College of Computer Science and Software Engineering,
    Shenzhen University, Shenzhen, China\\
    \textsuperscript{\rm 5}
    School of Artificial Intelligence,
    Shenzhen University, Shenzhen, China\\
    \textsuperscript{\rm 6}
    Software College,
    Northeastern University, Shenyang, China
}

\begin{document}
\maketitle

\begin{abstract}
Although self-supervised 3D anomaly detection assumes that acquiring high-precision point clouds is computationally expensive, in real manufacturing scenarios it is often feasible to collect a limited number of anomalous samples. Therefore, we study open-set supervised 3D anomaly detection, where the model is trained with only normal samples and a small number of known anomalous samples, aiming to identify unknown anomalies at test time.
We present Open-Industry, a high-quality industrial dataset containing 20 categories, each with five real anomaly types collected from production lines. We first adapt general open-set anomaly detection methods to accommodate 3D point cloud inputs better. Building upon this, we propose Open3D-AD, a point-cloud–oriented approach that leverages normal samples, simulated anomalies, and partially observed real anomalies to model the probability density distributions of normal and anomalous data. Then, we introduce a simple Correspondence Distributions Subsampling to reduce the overlap between normal and non-normal distributions, enabling stronger dual distributions modeling.
Based on these contributions, we establish a comprehensive benchmark and evaluate the proposed method extensively on Open-Industry as well as established datasets including Real3D-AD and Anomaly-ShapeNet. Benchmark results further reveal the potential of open-set supervised 3D anomaly detection. 
\end{abstract}

\begin{figure}
    \centering
    \includegraphics[width=0.95\linewidth]{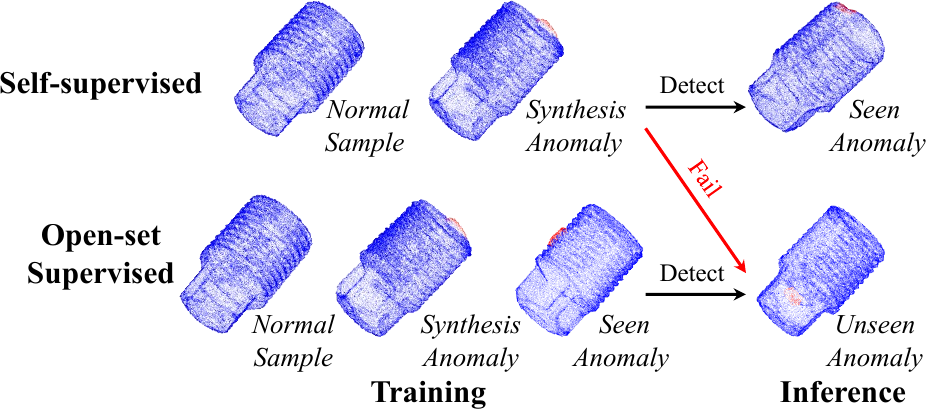}
    \caption{Comparison of 3D anomaly detection settings. Prior methods are unsupervised or rely on synthetic negatives without real negative-sample constraints. Consequently, the model lacks a clear optimization direction and has no reliable notion of the negative feature distribution, leading to weak anomaly awareness. We propose open-set supervised 3D anomaly detection, where the training set includes a few real anomalies from a subset of categories, enabling the model to learn a grounded notion of the negative feature distribution, while testing requires detecting unseen anomaly categories.}
    \label{first}
\end{figure}
\section{Introduction}
Point cloud anomaly detection is a 3D anomaly detection paradigm for high-resolution industrial quality inspection and has attracted increasing attention recently~\cite{liang20253dadsurvey,NEURIPS2024_9a263e23,bhunia2024looking}. Existing point cloud anomaly detection methods are predominantly unsupervised, or they rely on anomaly synthesis to create negative samples that assist feature representation learning~\cite{liu2023real3d,roth2022total,bergmann2023anomaly,ZAVRTANIK2024113}. This is mainly because point cloud acquisition is costly and difficult, and it is commonly assumed that collecting sufficiently diverse negative examples in industrial anomaly detection is impractical~\cite{li2025multi,guo2025iec3dad3ddatasetindustrial,bergmann2021mvtec}, as shown in Figure~\ref{first}. These models loosely treat anomalies as any structure that deviates from the normal distribution, while the model still struggles to learn an accurate notion of the true normal distribution~\cite{ZHAO2026108446,wang2023multimodal,lin2025commonality}. As a result, current feature embedding and feature reconstruction approaches lack clear optimization signals and anomaly awareness due to the absence of real negative-sample constraints~\cite{zhou2024r3dad,PO3AD}.

Therefore, it is crucial for real-world industrial applications that a model can leverage a small number of known negative examples to learn anomaly detection capability that generalizes across anomaly types. In this paper, we propose open-set supervised 3D anomaly detection, defined as the setting where the model is trained with normal reference samples and a few negative samples from only a subset of anomaly categories, yet is expected to detect unseen anomaly categories at test time.
Moreover, current point cloud datasets primarily focus on toy data or synthetic data, and lack simulation environments relevant to the industrial sector. To this end, we first introduce Open-Industry, a real-world dataset collected from production lines, consisting of common industrial components that comply with ISO-9001-2025 certification~\cite{tari2012benefits}. Each category is organized into normal references, seen anomalies, and unseen anomalies, aiming to faithfully emulate practical inspection scenarios in industrial manufacturing. In addition, Table~\ref{tab:3} summarises recent work on 3D anomaly detection datasets, and we introduced more related work in the \textit{Supplementary Materials}.

Building upon Open-Industry, we establish a dedicated benchmark to evaluate both representative open-set anomaly detection methods and 3D anomaly detection approaches from two perspectives: (1) how well general-purpose open-set anomaly detection methods transfer to point cloud data, and (2) how standard 3D anomaly detection methods perform under real production-line conditions. To address the above problems, our benchmark reveals that existing methods often exhibit weak point-level detection capability in point clouds, or struggle to effectively leverage limited negative samples because anomalous points are vastly outnumbered by normal points. To address these challenges, we propose Open3D-AD, an open-set supervised 3D anomaly detection method tailored for high-resolution industrial point clouds, which effectively tackles open-set anomaly detection in real-world manufacturing settings.
Our main contributions are summarized as follows:

\begin{enumerate}
    \item We propose the open-set supervised 3D anomaly detection task. To the best of our knowledge, we are the first to address this meaningful task of open-set  supervised anomaly detection on point clouds settings.
    \item We introduce Open-Industry, a real-world industrial point cloud dataset collected from production lines, comprising common industrial components that are systematically categorized into normal references, seen anomalies, and unseen anomalies. The dataset is designed to realistically simulate open-set supervised anomaly detection scenarios in industrial inspection.
    \item Based on Open-Industry, we establish a comprehensive benchmark for open-set supervised point cloud anomaly detection, enabling a unified evaluation of both general-purpose open-set anomaly detection methods and representative 3D anomaly detection approaches under realistic industrial settings.
    \item We propose Open3D-AD, a novel open-set supervised 3D anomaly detection framework that effectively leverages limited known anomalies to detect unseen anomaly categories in high-resolution industrial point clouds. On the proposed benchmark, Open3D-AD achieves state-of-the-art performance in terms of O-AUROC and P-AUROC for predicting both known and unknown anomalies.
\end{enumerate}

\begin{table}[t!]
    \centering
    \caption{Comparison of representative 3D anomaly detection datasets and learning settings. For fairness, only shared categories across datasets are compared. For the multimodal benchmarks MVTec 3D-AD and MulSen-AD, only anomalies observable in point clouds are considered. As IEC3D-AD is not open source, we contacted the author to obtain details.}
    \resizebox{\linewidth}{!}{
    \begin{tabular}{ccccc}
    \toprule
    \multicolumn{5}{c}{\textbf{Dataset Comparison}} \\
    \midrule
    Dataset & Class & Synthetic & Anomaly Types & Ratio \\
    \midrule
    MvTec3D-AD~\cite{bergmann2021mvtec} & 10    & Real  & 3   & $\sim$2\% \\
    Real3D-AD~\cite{liu2023real3d} & 12    & Real  & 2     & $\sim$3\% \\
    Anomaly-ShapeNet~\cite{Li_2024_CVPR} & 40    & Syn   & 4     & $\sim$2\% \\
    MulSen-AD~\cite{li2025multi} & 15    & Real  & 5     & $\sim$4\% \\
    IEC3D-AD\&IEC3D-IMAD~\cite{guo2025iec3dad3ddatasetindustrial} & 15 & Real & 2 & $\sim$1\% \\
    Real-IAD$^3$~\cite{Zhu_2025_CVPR} & 20 & Real & 2 & $\sim$1\% \\
    MiniShift~\cite{cheng2025towards} & 12    & Syn   & 4     & $\sim$1\% \\
    \hdashline
    \textbf{Open-Industry} & \textbf{20}    & \textbf{Real}  & \textbf{5}     & \textbf{$\sim$1\%} \\
    \midrule
    \midrule
    \multicolumn{5}{c}{\textbf{Task Comparison}} \\
    \midrule
    Benchmark & Setting & \multicolumn{3}{c}{Description} \\
    \midrule
    Real3D-AD~\cite{liu2023real3d} & \makecell{One-class\\ Self-supervised} & \multicolumn{3}{c}{No negative samples, detect known} \\
    MC3D-AD~\cite{ijcai2025p94} & \makecell{Multi-class\\Self-supervised} & \multicolumn{3}{c}{No negative samples, detect known} \\
    C3D-AD~\cite{lu2025c3dadcontinual3danomaly} & \makecell{Continual-Learning\\Self-supervised} & \multicolumn{3}{c}{No negative samples, detect unknown } \\
    \hdashline
    \textbf{Open-Industry} & \makecell{\textbf{Open-set}\\\textbf{Supervised}} & \multicolumn{3}{c}{\textbf{Has negative samples, detect unknown }} \\
    \bottomrule
    \end{tabular}%
    }
    \label{tab:3}
\end{table}

\section{Related Work}

\subsection{3D Anomaly Detection}
Existing 3D anomaly detection methods commonly characterize normal geometry through feature matching or geometric restoration. Memory-based approaches represent normality using reference features and score test samples by their deviations from these references. Originally developed for image anomaly detection, PatchCore~\cite{roth2022total} constructs a coreset-subsampled memory bank of normal patch features for nearest-neighbor scoring. BTF~\cite{horwitz2023back} demonstrates the effectiveness of FPFH-based geometric descriptors and combines them with appearance features, while M3DM~\cite{wang2023multimodal} combines RGB and point-cloud representations through feature fusion and decision fusion with multiple memory banks. For point-cloud inputs, ISMP~\cite{liang2025look} complements local point features with internal spatial representations and filters redundant information before memory-based scoring. Simple3D~\cite{cheng2025towards} combines multi-scale geometric descriptors with local spatial aggregation and detects anomalies by measuring deviations from normal prototypes. Beyond feature matching, geometric restoration and offset prediction provide another route to detecting abnormal structures. R3D-AD~\cite{zhou2024r3dad} uses diffusion-based reconstruction with step-wise point displacement and introduces Patch-Gen to synthesize geometric defects for training. PO3AD~\cite{PO3AD} learns point-wise offset prediction using normal-vector-guided synthetic anomalies and uses predicted displacement magnitudes as anomaly scores. These approaches obtain anomaly-related training signals from synthesized defects. Our framework adopts the multi-scale geometric representation used in Simple3D and the Norm-AS synthesis strategy from PO3AD as its feature extraction and anomaly augmentation components, respectively.

\subsection{Open-Set Supervised Anomaly Detection}

Limited anomaly supervision provides information about abnormal patterns that can complement normal-reference modeling. DevNet~\cite{pang2019deep} directly learns anomaly scores from a small set of labeled anomalies and unlabeled data. Its deviation loss uses a Gaussian reference distribution to encourage labeled anomalies to receive scores that deviate from normal behavior. Open-set supervised anomaly detection further considers incomplete coverage of anomaly categories during training, requiring a detector to recognize unseen anomaly types. DRA~\cite{ding2022catching} addresses this setting by learning disentangled representations of seen, pseudo, and latent residual abnormalities, with the latter two supporting generalization to unseen anomalies.

We study this setting for industrial point clouds and adapt DevNet and DRA using point-cloud encoders pretrained with Point-MAE~\cite{pang2022masked} or Point-BERT~\cite{yu2021pointbert}. Open3D-AD combines normal references, observed real anomalies, and synthesized anomalies in separate normal and anomalous memory banks. It uses the normal support to refine the anomalous support and combines distances to both for anomaly scoring. This formulation incorporates limited real-anomaly supervision into geometric feature matching for detecting unseen defect types.

\section{Open-Industry Benchmark}
To address the need for high-fidelity, real-world industrial scenarios and diverse anomaly patterns~\cite{guo2025point,11361169,wang2024real,costanzino2025sim3d,bonfiglioli2022eyecandies,wang2024real}, we construct \emph{Open-Industry}, a large-scale 3D anomaly detection benchmark collected directly from real industrial production lines. Unlike existing benchmarks built from toy objects or synthetic generation, our benchmark is derived from practical machining environments and preserves authentic manufacturing traces, surface textures, and process-induced defects. 

\subsection{Benchmark Construction}
To ensure the reliability of benchmark construction, all samples are collected directly from real industrial machining production lines. The benchmark covers 20 categories of industrial equipment components, each containing both normal and anomalous samples. The anomalous subset includes five representative process-induced defect types, namely \emph{convex}, \emph{concave}, \emph{scratch}, \emph{scar}, and \emph{deformation}. To obtain accurate point cloud measurements, we build a structured-light acquisition system consisting of two CCD cameras, a structured-light projector, and a $360^{\circ}$ automatic turntable, as summarized in the \textit{Supplementary Materials}. During data construction, all samples are processed through a complete pipeline including system calibration, multi-view registration, denoising, resampling, and fine-grained defect annotation. Specifically, the acquisition system is first calibrated to estimate intrinsic and extrinsic parameters for accurate stereo reconstruction; the reconstructed point clouds from multiple viewpoints are then aligned into a unified coordinate system through coarse-to-fine registration; the resulting point clouds are further denoised and resampled to remove outliers while preserving geometric structure; finally, anomalous regions are manually annotated and reviewed to ensure reliable defect localization and annotation quality. More technical details of the calibration and processing procedures are provided in the \textit{Supplementary Materials}.


\subsection{Data Description}
The Open-Industry benchmark contains 2,400 point clouds from 20 categories of real industrial components, including washers, pins, screws, and nuts. Each category comprises 100 normal and 60 anomalous samples, yielding 1,500 normal and 900 anomalous samples in total. The anomalies cover five process-induced defect types: \emph{convex}, \emph{concave}, \emph{scratch}, \emph{scar}, and \emph{deformation}. A key challenge is the extreme sparsity of defects, with anomaly point ratios ranging from 0.51\% to 2.64\% and averaging only 1.20\%. Moreover, scratches and scars are highly localized, while convex, concave, and deformation defects may overlap with normal functional structures, leading to weak geometric contrast and ambiguous boundaries. All samples are manually inspected to ensure accurate alignment, reliable annotations, and consistent quality. After preprocessing, each point cloud contains approximately 90K points on average. Overall, Open-Industry provides a realistic benchmark for evaluating the generalization of 3D anomaly detection methods in practical industrial scenarios.

\subsection{Task Setting}
We consider open-set supervised 3D anomaly detection, where the model is trained with abundant normal samples and a limited number of labeled anomalies from selected seen categories, and is required to detect previously unseen anomaly categories at test time~\cite{ding2022catching}. To ensure fair comparison and systematically evaluate the effect of supervision strength, we construct two training settings with different numbers of labeled anomalies. Specifically, for each dataset, a subset of anomaly categories is designated as seen while the remaining categories are treated as unseen, and two different quantities of anomalous samples are drawn from the seen categories for supervision. The number of labeled anomalies serves as a proxy for supervision strength, allowing us to analyze its impact on detection performance and generalization to unseen anomalies.

\begin{figure*}[!t]
    \centering
\includegraphics[width=0.8\linewidth]{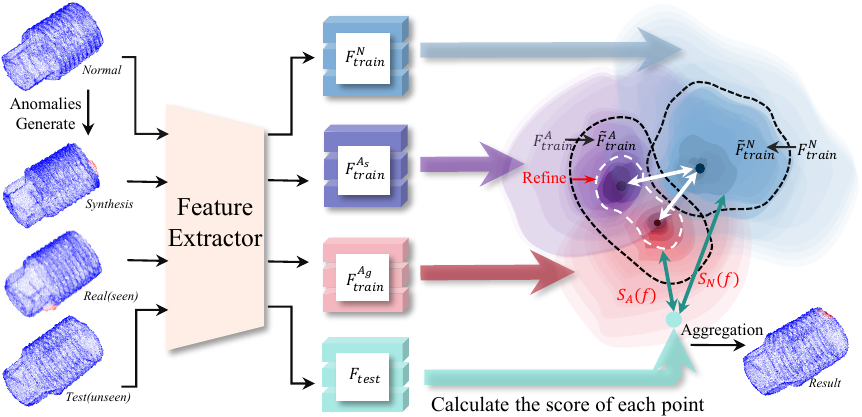}
    \caption{Workflow of our Open3D-AD. It consists of three steps: (i) anomaly synthesis and point-wise feature encoding to form normal/anomalous mixture representations, (ii) greedily subsample the normal support and use it to refine the anomalous support via deconfounding, and (iii) dual-distribution aggregated scoring for anomaly detection.}
    \label{fig:pipeline}
\end{figure*}

\section{Methodology}
\subsection{Problem Statement}
Let $\mathcal{X}$ denote the space of point-cloud samples. We are given a normal set $\mathcal{P}^N\subset\mathcal{X}$ and an anomalous set $\mathcal{P}^A\subset\mathcal{X}$, with $\mathcal{P}^N\cap\mathcal{P}^A=\emptyset$. In open-set supervised 3D anomaly detection, the anomalous set is further decomposed into two disjoint subsets, the seen anomalies and the unseen anomalies, i.e., $\mathcal{P}^A=\mathcal{P}^{A_s}\cup\mathcal{P}^{A_u}$ and $\mathcal{P}^{A_s}\cap\mathcal{P}^{A_u}=\emptyset$. During training, only abundant normal samples and a small number of seen anomalous samples are accessible, namely $\mathcal{P}_{train}=(\mathcal{P}^N_{train}\cup\mathcal{P}^{A_s}_{train}) \subseteq (\mathcal{P}^N\cup\mathcal{P}^{A_s})$. At test time, the evaluation set consists of normal samples and unseen anomalous samples, i.e., $\mathcal{P}_{test}=(\mathcal{P}^N_{test}\cup\mathcal{P}^{A_u}_{test})\subseteq (\mathcal{P}^N\cup\mathcal{P}^{A_u})$. The goal is to learn an anomaly scoring function $f_\theta:\mathcal{X}\rightarrow\mathbb{R}$ that assigns low scores to normal samples and high scores to unseen anomalies, thereby enabling generalizable detection of unseen anomalous patterns.

\subsection{Overview}
Given the limited size of the anomalous training set $\mathcal{P}^{A_s}_{train}$, the pipeline first introduces an anomaly synthesis stage to augment the observed anomalies, producing a simulated anomaly set $\hat{\mathcal{P}}^{A_g}_{train}$ that enriches the anomalous distribution.
Subsequently, the framework constructs a unified feature space via a geometric operator $\mathcal{F}$. Specifically, both the union of seen and simulated anomalies, i.e., $\mathcal{P}^{A_s}_{train}\cup \hat{\mathcal{P}}^{A_g}_{train}$, and the normal training set $\mathcal{P}^{N}_{train}$ are encoded within this space, enabling an explicit characterization of the normal and anomalous distributions.
At inference time, for any test sample $x\in\mathcal{P}_{test}$, the pipeline evaluates its relative distances to the normal and anomalous distributions in the induced feature space and jointly determines the final anomaly score $f_\theta$. The overall workflow is illustrated in Figure~\ref{fig:pipeline}.

\subsection{Feature Distributions Representations}
This section details the feature extraction process for open-set supervised 3D anomaly detection.
Due to the limited number of seen anomalous samples $\mathcal{P}^{A_s}_{train}$, 
we adopt an existing point-cloud anomaly synthesis method for data augmentation. 
Specifically, Norm-AS~\cite{PO3AD} is employed to generate simulated anomalies 
by applying local geometric perturbations to normal point clouds. 
For a normal sample $x \in \mathcal{P}^{N}_{train}$, the synthesized anomalous sample is defined as
\begin{equation}
\hat{x} = \mathcal{T}(x),
\end{equation}
where $\mathcal{T}(\cdot)$ denotes the anomaly synthesis operation. 
The generated anomalous samples constitute the simulated anomaly set 
$\mathcal{P}^{A_g}_{train} = \{ \mathcal{T}(x) \mid x \in \mathcal{P}^{N}_{train} \}$.
Thus, the augmented training set is defined as
\begin{equation}
\hat{\mathcal{P}}_{train} 
= \left( \mathcal{P}^{N}_{train} 
\cup \mathcal{P}^{A_s}_{train} 
\cup \mathcal{P}^{A_g}_{train} \right).
\end{equation}

For feature extraction, following previous point-cloud anomaly detection methods, 
we adopt multiple fast point feature histogram (multi-FPFH)~\cite{rusu2009fast,cheng2025towards} 
as the feature extractor, denoted as $f_\phi$. 
For any point cloud $\mathcal{P} \in \mathcal{X}$, its feature representation is given by
\begin{equation}
F = f_\phi(\mathcal{P}), 
\quad F \in \mathbb{R}^{G \times C_1},
\end{equation}
where $F$ denotes the extracted feature representation, 
with $G$ representing the number of sampled center points using farthest point sampling~(FPS)~\cite{qi2017pointnet,qi2017pointnetplus} and $C_1$ denoting the feature channel dimension. 
Accordingly,
\begin{equation}
\begin{aligned}
F_{train} &= \{ f_\phi(x) \mid x \in \hat{\mathcal{P}}_{train} \}, \\
F_{test}  &= \{ f_\phi(x) \mid x \in \mathcal{P}_{test} \},
\end{aligned}
\end{equation}
The training features can be further decomposed as
\begin{equation}
\begin{aligned}
F_{train}^{N} &= \{ f_\phi(x) \mid x \in \mathcal{P}^{N}_{train} \}, \\
F_{train}^{A} &= \{ f_\phi(x) \mid x \in \left( \mathcal{P}^{A_s}_{train} \cup \mathcal{P}^{A_g}_{train} \right) \}.
\end{aligned}
\end{equation}

To provide a probabilistic interpretation, we view the extracted features as drawn from
mixture distributions composed of normal and anomalous components. Specifically, the
training features follow
\begin{equation}
p_{\text{train}}(F)=\pi_N^{\text{train}} p_N(F) + \pi_A^{\text{train}} p_A(F),
\end{equation}
where $p_N(F)$ and $p_A(F)$ denote the normal and (observed) anomalous feature
distributions induced by $\mathcal{P}^{N}_{\text{train}}$ and
$\mathcal{P}^{A_s}_{\text{train}} \cup \mathcal{P}^{A_g}_{\text{train}}$, respectively.
$\pi_N^{\text{train}}$ and $\pi_A^{\text{train}}$ are the corresponding mixture proportions
with $\pi_N^{\text{train}}+\pi_A^{\text{train}}=1$.
Accordingly, the evaluation set contains both normal samples and unseen anomalies as,
\begin{equation}
p_{\text{test}}(F)=\pi_N^{\text{test}} p_N(F) + \pi_A^{\text{test}} p_{A_u}(F),
\end{equation}
where $p_{A_u}(F)$ denotes the (unknown) feature distribution of unseen anomalies, and
$\pi_N^{\text{test}}+\pi_A^{\text{test}}=1$.
Note that the mixture proportions $\pi^{\text{train}}$ and $\pi^{\text{test}}$ are introduced for interpretation; our method does not require explicitly estimating them.


\subsection{Correspondence Distributions Subsampling}
While coreset subsampling has been widely demonstrated to effectively represent the support of an underlying distribution~\cite{roth2022total}, its extension to dual-distribution modeling in open-set scenarios remains to be further explored. To this end, we define a greedy subsampling operator $\mathcal{S}(\cdot, k)$ that extracts an optimal discrete support of cardinality $k$ from a given feature set. This operator is employed to refine the empirical representations of $p_N(F)$ and $p_A(F)$ while enhancing their discriminative boundary.

Specifically, we first derive a compact discrete support $\tilde{F}_{train}^{N}$ for the normal distribution $p_N(F)$ as
\begin{equation}
\begin{aligned}
    \tilde{F}_{train}^{N} &= \mathcal{S}(F_{train}^{N}, 
    \mathcal{N}), \\\tilde{F}_{train}^{N} &\sim p_N(F),
\end{aligned}
\end{equation}
where $\mathcal{N}$ denotes the target number of representative features. To refine the anomalous distribution $p_A(F)$, we initially sample a candidate support $\check{F}_{train}^{A} = \mathcal{S}(F_{train}^{A}, 2\mathcal{N})$, where $\check{F}_{train}^{A} \sim p_A(F)$. To minimize the distributional overlap between $p_N(F)$ and $p_A(F)$, we identify the components in $\check{F}_{train}^{A}$ that are most likely to be confounded with the normal manifold. For each $f \in \tilde{F}_{train}^{N}$, its corresponding neighbor in the anomalous candidate set is defined as
\begin{equation}
F_{cor} = \left\{ \arg\min_{g \in \check{F}_{train}^{A}} \| f - g \|_2 \;\Big|\; f \in \tilde{F}_{train}^{N} \right\},
\end{equation}
From a probabilistic perspective, $F_{cor}$ represents the subset of $\check{F}_{train}^{A}$ that maximizes the local density overlap with the normal support. Thus, by removing these confounded components, the final refined anomalous support is given by
\begin{equation}
\begin{aligned}
    \tilde{F}_{train}^{A},  &= \check{F}_{train}^{A} \setminus F_{cor}, \\
    |\tilde{F}_{train}^{A}| &= \mathcal{N},
\end{aligned}
\end{equation}
Accordingly, $(\tilde{F}_{train}^{N}, \tilde{F}_{train}^{A})$ constitutes a balanced and well-separated empirical basis for $p_N(F)$ and $p_A(F)$, facilitating the subsequent optimization of the anomaly scoring function $f_\theta$.

\subsection{Anomaly Score Calculation}
The anomaly score is determined by evaluating the proximity of test features to the refined empirical supports $(\tilde{F}_{train}^{N}, \tilde{F}_{train}^{A})$. For any feature $f \in F_{test}$, let $g^* \in \tilde{F}_{train}^{N}$ and $h^* \in \tilde{F}_{train}^{A}$ be its nearest neighbors. To enhance robustness, we adopt an importance reweighting method~\cite{liu2015classification} and define a distance operator $\mathcal{D}(f, q, \tilde{F})$ as:
\begin{equation}
\mathcal{D}(f, q, \tilde{F}) = \left( 1 - \frac{\exp \| f - q \|_2}{\sum_{m \in \mathcal{N}_{\mathcal{K}}(q)} \exp \| f - m \|_2} \right) \cdot \| f - q \|_2,
\end{equation}
where $\mathcal{N}_{\mathcal{K}}(q) \subset \tilde{F}$ is the set of $\mathcal{K}$-nearest neighbors ($\mathcal{K}$-NN)~\cite{kramer2013k} of $q$. Accordingly, the distribution-aware distances are $s_N(f) = \mathcal{D}(f, g^*, \tilde{F}_{train}^{N})$ and $s_A(f) = \mathcal{D}(f, h^*, \tilde{F}_{train}^{A})$. The local score $\alpha(f)$ is derived as:
\begin{equation}
\alpha(f) = (1 - \gamma \cdot s_A(f)) \cdot s_N(f),
\end{equation}
where $\gamma$ scales the influence of the anomalous distribution. For a point-cloud sample $x$, the point-level score for each constituent point $p_i$ is $\mathcal{S}_{point}(p_i) = \alpha(f_i)$. The final anomaly scoring function $f_\theta:\mathcal{X}\rightarrow\mathbb{R}$ is defined as the maximum point-level score:
\begin{equation}
f_\theta(x) = \max_{p_i \in x} \mathcal{S}_{point}(p_i).
\end{equation}

\begin{table*}[!t]
    \centering
    \caption{Object-AUROC/P-AUROC results under the open-set protocol, where models are trained using five or ten known anomaly classes and evaluated on detecting the remaining overall anomaly categories in the Open-Industry dataset. Results are reported as mean $\pm$ standard deviation, with object-level and point-level metrics separated by ``/''.}
    \label{o-open}
    \resizebox{\textwidth}{!}{%
    \begin{minipage}{1.66\textwidth}
    \centering
    \setlength{\tabcolsep}{3pt}
    \renewcommand{\arraystretch}{0.95}
    \footnotesize
    \begin{tabular*}{\linewidth}{@{\extracolsep{\fill}}clcccccccc@{}}
\toprule
\multicolumn{10}{c}{\textbf{(a) O-AUROC/P-AUROC}} \\
\midrule
Setting & Method & Square Taper Washer & Split Pin & Lock Nut & Threaded Nail & L-shaped Screw & Circlip & Internal Tooth Lock Washer & Double Lug Washer \\
\midrule
\multirow{5}{*}{\makecell{Five\\samples}} & DevNet$_{\mathrm{pm}}$ & 56.96$_{\pm3.31}$/50.94$_{\pm2.53}$ & 59.65$_{\pm1.71}$/48.10$_{\pm2.57}$ & 59.43$_{\pm3.62}$/\underline{51.18}$_{\pm3.24}$ & 55.67$_{\pm3.88}$/48.16$_{\pm1.42}$ & 55.13$_{\pm7.94}$/51.22$_{\pm3.07}$ & 56.61$_{\pm4.24}$/47.62$_{\pm2.14}$ & 49.83$_{\pm2.31}$/50.28$_{\pm1.17}$ & 52.01$_{\pm5.03}$/51.00$_{\pm5.43}$ \\
 & DevNet$_{\mathrm{pb}}$ & \underline{68.36}$_{\pm1.97}$/51.31$_{\pm3.29}$ & 67.93$_{\pm3.76}$/49.20$_{\pm1.49}$ & \underline{65.69}$_{\pm2.49}$/48.94$_{\pm2.39}$ & 57.93$_{\pm3.07}$/49.60$_{\pm3.42}$ & 61.31$_{\pm1.53}$/50.88$_{\pm1.76}$ & 61.15$_{\pm3.24}$/49.56$_{\pm2.45}$ & \underline{60.95}$_{\pm3.25}$/50.41$_{\pm0.90}$ & 67.46$_{\pm4.31}$/49.20$_{\pm4.26}$ \\
 & DRA$_{\mathrm{pm}}$ & 66.81$_{\pm1.87}$/\underline{56.02}$_{\pm3.36}$ & 65.85$_{\pm1.41}$/\underline{58.75}$_{\pm3.19}$ & 49.77$_{\pm3.88}$/49.40$_{\pm1.90}$ & \underline{64.43}$_{\pm3.06}$/\underline{56.98}$_{\pm8.33}$ & \underline{63.83}$_{\pm1.81}$/49.62$_{\pm1.06}$ & 58.30$_{\pm3.60}$/\underline{54.29}$_{\pm3.71}$ & 54.80$_{\pm1.82}$/\underline{56.12}$_{\pm4.13}$ & 67.60$_{\pm2.82}$/\underline{66.06}$_{\pm3.86}$ \\
 & DRA$_{\mathrm{pb}}$ & 67.63$_{\pm3.76}$/51.39$_{\pm1.90}$ & \underline{73.54}$_{\pm4.27}$/51.61$_{\pm4.05}$ & 63.19$_{\pm4.95}$/50.22$_{\pm1.76}$ & 59.34$_{\pm4.74}$/51.40$_{\pm6.61}$ & 61.46$_{\pm6.18}$/\underline{51.75}$_{\pm2.76}$ & \underline{71.15}$_{\pm4.97}$/48.88$_{\pm3.16}$ & 59.84$_{\pm5.46}$/51.16$_{\pm2.30}$ & \underline{70.72}$_{\pm6.33}$/49.58$_{\pm6.05}$ \\
 & Ours & \textbf{89.60}$_{\pm1.46}$/\textbf{71.85}$_{\pm1.68}$ & \textbf{82.61}$_{\pm1.20}$/\textbf{75.57}$_{\pm3.46}$ & \textbf{78.47}$_{\pm6.87}$/\textbf{69.48}$_{\pm1.97}$ & \textbf{84.51}$_{\pm3.92}$/\textbf{78.76}$_{\pm2.04}$ & \textbf{90.52}$_{\pm2.40}$/\textbf{73.70}$_{\pm3.39}$ & \textbf{90.09}$_{\pm2.64}$/\textbf{70.34}$_{\pm4.41}$ & \textbf{79.23}$_{\pm6.12}$/\textbf{64.75}$_{\pm2.92}$ & \textbf{90.13}$_{\pm1.50}$/\textbf{71.67}$_{\pm1.79}$ \\
\midrule
\multirow{5}{*}{\makecell{Ten\\samples}} & DevNet$_{\mathrm{pm}}$ & 58.67$_{\pm4.91}$/49.95$_{\pm2.19}$ & 56.85$_{\pm5.89}$/51.10$_{\pm4.38}$ & 62.18$_{\pm3.97}$/49.28$_{\pm1.98}$ & 55.96$_{\pm3.68}$/52.94$_{\pm3.75}$ & 54.82$_{\pm3.80}$/48.99$_{\pm2.08}$ & 59.07$_{\pm2.10}$/49.75$_{\pm2.80}$ & 59.58$_{\pm4.63}$/51.09$_{\pm1.99}$ & 55.14$_{\pm2.04}$/52.24$_{\pm6.55}$ \\
 & DevNet$_{\mathrm{pb}}$ & 67.81$_{\pm1.56}$/50.88$_{\pm3.43}$ & 68.62$_{\pm3.91}$/48.83$_{\pm0.83}$ & 63.62$_{\pm2.72}$/49.37$_{\pm4.30}$ & 58.03$_{\pm3.06}$/49.26$_{\pm3.35}$ & 61.87$_{\pm1.93}$/51.12$_{\pm1.03}$ & 61.85$_{\pm4.15}$/49.21$_{\pm3.07}$ & 58.83$_{\pm3.37}$/49.80$_{\pm1.32}$ & 68.22$_{\pm4.08}$/49.58$_{\pm4.59}$ \\
 & DRA$_{\mathrm{pm}}$ & \underline{69.71}$_{\pm1.27}$/\underline{57.85}$_{\pm4.30}$ & \underline{78.08}$_{\pm2.57}$/\underline{59.38}$_{\pm1.81}$ & 70.06$_{\pm5.06}$/\underline{50.72}$_{\pm3.18}$ & \underline{72.59}$_{\pm2.60}$/\underline{62.44}$_{\pm4.26}$ & 61.52$_{\pm1.86}$/\underline{52.95}$_{\pm3.99}$ & 74.13$_{\pm3.06}$/\underline{53.63}$_{\pm1.76}$ & \underline{68.05}$_{\pm1.82}$/\underline{59.73}$_{\pm3.10}$ & 65.74$_{\pm1.95}$/\underline{62.36}$_{\pm6.10}$ \\
 & DRA$_{\mathrm{pb}}$ & 69.51$_{\pm1.94}$/51.70$_{\pm4.88}$ & 70.53$_{\pm2.73}$/50.56$_{\pm1.07}$ & \underline{71.51}$_{\pm2.70}$/49.17$_{\pm3.15}$ & 61.33$_{\pm2.40}$/58.74$_{\pm2.71}$ & \underline{66.95}$_{\pm5.61}$/50.79$_{\pm1.97}$ & \underline{76.49}$_{\pm3.61}$/48.03$_{\pm3.07}$ & 61.15$_{\pm4.89}$/51.81$_{\pm1.92}$ & \underline{68.42}$_{\pm3.68}$/57.44$_{\pm4.86}$ \\
 & Ours & \textbf{88.86}$_{\pm1.47}$/\textbf{70.25}$_{\pm4.61}$ & \textbf{86.72}$_{\pm1.78}$/\textbf{84.82}$_{\pm1.95}$ & \textbf{73.58}$_{\pm6.88}$/\textbf{73.21}$_{\pm4.40}$ & \textbf{83.30}$_{\pm2.76}$/\textbf{84.25}$_{\pm1.63}$ & \textbf{88.15}$_{\pm1.73}$/\textbf{85.34}$_{\pm0.97}$ & \textbf{86.81}$_{\pm3.36}$/\textbf{87.00}$_{\pm1.38}$ & \textbf{78.75}$_{\pm4.40}$/\textbf{68.31}$_{\pm2.60}$ & \textbf{89.64}$_{\pm1.22}$/\textbf{69.61}$_{\pm3.93}$ \\
\bottomrule
\noalign{\vskip 0.6em}
\toprule
Setting & Method & Square Plug & S-hook & External Tooth Lock Washer & Hex Head Plug & Eyelet Self-tapping Screw & Heavy Washer & Locking Tab Washer & Average \\
\midrule
\multirow{5}{*}{\makecell{Five\\samples}} & DevNet$_{\mathrm{pm}}$ & 63.03$_{\pm4.57}$/51.25$_{\pm3.17}$ & 53.26$_{\pm4.15}$/49.59$_{\pm2.17}$ & 58.02$_{\pm7.12}$/\underline{53.08}$_{\pm2.05}$ & 61.47$_{\pm4.83}$/51.36$_{\pm3.04}$ & 56.61$_{\pm4.94}$/\underline{50.85}$_{\pm3.04}$ & 62.58$_{\pm3.40}$/50.31$_{\pm2.24}$ & 58.07$_{\pm4.37}$/\underline{54.42}$_{\pm1.19}$ & 57.22$_{\pm1.42}$/50.62$_{\pm0.77}$ \\
 & DevNet$_{\mathrm{pb}}$ & 67.25$_{\pm2.53}$/50.99$_{\pm2.98}$ & 62.70$_{\pm2.22}$/49.87$_{\pm2.00}$ & 63.05$_{\pm2.00}$/52.16$_{\pm4.80}$ & 70.15$_{\pm1.92}$/\underline{56.04}$_{\pm3.43}$ & 60.66$_{\pm2.99}$/49.56$_{\pm1.49}$ & 66.15$_{\pm3.98}$/50.04$_{\pm3.99}$ & 66.97$_{\pm1.31}$/49.66$_{\pm1.63}$ & 64.51$_{\pm1.15}$/50.49$_{\pm0.69}$ \\
 & DRA$_{\mathrm{pm}}$ & 56.12$_{\pm1.22}$/51.65$_{\pm2.88}$ & \underline{71.43}$_{\pm4.46}$/\underline{55.51}$_{\pm6.76}$ & \underline{71.18}$_{\pm5.19}$/52.13$_{\pm5.40}$ & \underline{75.33}$_{\pm5.37}$/49.91$_{\pm2.59}$ & 64.33$_{\pm6.43}$/49.69$_{\pm4.35}$ & 65.68$_{\pm6.24}$/\underline{57.90}$_{\pm4.31}$ & 68.12$_{\pm6.50}$/51.29$_{\pm3.91}$ & 64.24$_{\pm2.21}$/\underline{54.35}$_{\pm1.47}$ \\
 & DRA$_{\mathrm{pb}}$ & \underline{70.69}$_{\pm4.06}$/\underline{51.89}$_{\pm3.55}$ & 64.36$_{\pm6.31}$/53.76$_{\pm5.69}$ & 66.38$_{\pm6.26}$/49.71$_{\pm4.61}$ & 74.20$_{\pm4.04}$/53.77$_{\pm2.36}$ & \underline{65.13}$_{\pm5.43}$/48.86$_{\pm4.57}$ & \underline{68.18}$_{\pm4.66}$/52.71$_{\pm6.11}$ & \underline{70.37}$_{\pm4.92}$/52.42$_{\pm4.49}$ & \underline{67.08}$_{\pm3.22}$/51.27$_{\pm0.86}$ \\
 & Ours & \textbf{76.47}$_{\pm1.98}$/\textbf{64.89}$_{\pm2.49}$ & \textbf{94.44}$_{\pm0.89}$/\textbf{74.21}$_{\pm2.08}$ & \textbf{80.68}$_{\pm5.79}$/\textbf{70.89}$_{\pm2.02}$ & \textbf{82.95}$_{\pm4.16}$/\textbf{60.44}$_{\pm3.43}$ & \textbf{85.85}$_{\pm1.44}$/\textbf{56.97}$_{\pm1.21}$ & \textbf{77.08}$_{\pm2.21}$/\textbf{72.11}$_{\pm2.67}$ & \textbf{83.23}$_{\pm2.38}$/\textbf{82.16}$_{\pm1.64}$ & \textbf{84.39}$_{\pm0.44}$/\textbf{70.52}$_{\pm0.56}$ \\
\midrule
\multirow{5}{*}{\makecell{Ten\\samples}} & DevNet$_{\mathrm{pm}}$ & 68.32$_{\pm2.88}$/51.04$_{\pm2.58}$ & 56.32$_{\pm1.89}$/51.66$_{\pm3.17}$ & 55.84$_{\pm5.92}$/50.80$_{\pm2.76}$ & 63.52$_{\pm3.62}$/52.18$_{\pm3.24}$ & 51.97$_{\pm1.81}$/\underline{51.55}$_{\pm3.66}$ & 62.05$_{\pm2.96}$/52.77$_{\pm5.89}$ & 59.81$_{\pm4.01}$/\underline{53.88}$_{\pm2.67}$ & 58.67$_{\pm1.64}$/51.28$_{\pm1.93}$ \\
 & DevNet$_{\mathrm{pb}}$ & 68.06$_{\pm2.24}$/50.09$_{\pm3.54}$ & 62.80$_{\pm1.47}$/49.52$_{\pm1.85}$ & 63.09$_{\pm3.49}$/51.91$_{\pm4.88}$ & 71.67$_{\pm2.38}$/\underline{54.91}$_{\pm3.49}$ & 61.98$_{\pm2.30}$/49.53$_{\pm1.49}$ & 65.12$_{\pm4.34}$/50.47$_{\pm3.90}$ & 69.56$_{\pm3.95}$/50.85$_{\pm2.63}$ & 64.74$_{\pm1.07}$/50.36$_{\pm1.23}$ \\
 & DRA$_{\mathrm{pm}}$ & \underline{72.37}$_{\pm3.49}$/\underline{53.79}$_{\pm4.64}$ & 64.90$_{\pm2.60}$/\underline{59.02}$_{\pm2.94}$ & \underline{74.03}$_{\pm0.96}$/\underline{55.18}$_{\pm6.29}$ & \underline{74.73}$_{\pm2.67}$/50.80$_{\pm1.13}$ & \underline{73.79}$_{\pm3.83}$/51.47$_{\pm5.22}$ & \textbf{73.14}$_{\pm6.65}$/\underline{54.06}$_{\pm8.27}$ & 68.83$_{\pm6.95}$/52.77$_{\pm2.65}$ & \underline{70.78}$_{\pm0.80}$/\underline{55.74}$_{\pm1.31}$ \\
 & DRA$_{\mathrm{pb}}$ & 71.28$_{\pm2.51}$/50.81$_{\pm2.58}$ & \underline{64.95}$_{\pm3.33}$/53.37$_{\pm8.28}$ & 66.04$_{\pm3.00}$/49.10$_{\pm5.21}$ & 73.31$_{\pm1.34}$/53.10$_{\pm4.06}$ & 64.70$_{\pm3.18}$/50.52$_{\pm2.34}$ & 65.28$_{\pm2.94}$/52.80$_{\pm4.63}$ & \underline{75.71}$_{\pm1.36}$/52.52$_{\pm3.43}$ & 68.48$_{\pm0.57}$/52.03$_{\pm1.69}$ \\
 & Ours & \textbf{76.01}$_{\pm4.05}$/\textbf{68.09}$_{\pm1.02}$ & \textbf{96.22}$_{\pm1.04}$/\textbf{87.32}$_{\pm2.71}$ & \textbf{80.83}$_{\pm1.61}$/\textbf{81.61}$_{\pm2.37}$ & \textbf{75.46}$_{\pm4.87}$/\textbf{62.89}$_{\pm2.46}$ & \textbf{86.34}$_{\pm2.83}$/\textbf{60.38}$_{\pm2.01}$ & \underline{69.36}$_{\pm6.33}$/\textbf{75.01}$_{\pm2.44}$ & \textbf{81.12}$_{\pm1.29}$/\textbf{86.21}$_{\pm1.92}$ & \textbf{82.74}$_{\pm0.79}$/\textbf{76.29}$_{\pm0.77}$ \\
\bottomrule
\end{tabular*}
    \end{minipage}%
    }
\end{table*}

\begin{table*}[!t]
    \centering
    \caption{Object-AUROC/P-AUROC results under the open-set protocol, where models are trained using two or four known anomaly classes and evaluated on detecting the remaining overall anomaly categories in the Real3D-AD dataset. Results are reported as mean $\pm$ standard deviation, with object-level and point-level metrics separated by ``/''.}
    \label{o-real3dad}
    \setlength{\tabcolsep}{2pt}
    \renewcommand{\arraystretch}{0.95}
    \resizebox{\textwidth}{!}{%
    \begin{tabular}{c||ccccc|ccccc}
        \hline
        \raisebox{-1.45ex}[0pt][0pt]{Dataset} & & & \multicolumn{1}{c}{Two Sample per class} & & & & & \multicolumn{1}{c}{Four Sample per class} & & \\
        \cline{2-11}
        & DevNet$_{\mathrm{pm}}$ & DevNet$_{\mathrm{pb}}$ & DRA$_{\mathrm{pm}}$ & DRA$_{\mathrm{pb}}$ & Ours & DevNet$_{\mathrm{pm}}$ & DevNet$_{\mathrm{pb}}$ & DRA$_{\mathrm{pm}}$ & DRA$_{\mathrm{pb}}$ & Ours \\
        \hline
        Airplane & \underline{50.90}$_{\pm3.37}$/56.36$_{\pm13.78}$ & 48.63$_{\pm1.60}$/35.05$_{\pm9.82}$ & 50.27$_{\pm0.00}$/\underline{69.67}$_{\pm0.00}$ & 46.66$_{\pm4.06}$/43.39$_{\pm9.98}$ & \textbf{57.85}$_{\pm5.60}$/\textbf{87.98}$_{\pm0.93}$ & 48.88$_{\pm2.33}$/57.63$_{\pm10.15}$ & 49.36$_{\pm2.13}$/29.05$_{\pm2.54}$ & \underline{50.18}$_{\pm2.37}$/\underline{57.95}$_{\pm10.97}$ & 48.47$_{\pm3.14}$/37.28$_{\pm1.89}$ & \textbf{56.14}$_{\pm6.93}$/\textbf{88.43}$_{\pm0.48}$ \\
        Candybar & 49.34$_{\pm2.86}$/58.24$_{\pm9.22}$ & \underline{53.47}$_{\pm3.29}$/57.76$_{\pm4.04}$ & 49.11$_{\pm2.98}$/\underline{60.89}$_{\pm5.82}$ & 48.54$_{\pm1.85}$/55.19$_{\pm7.37}$ & \textbf{83.93}$_{\pm6.54}$/\textbf{94.16}$_{\pm0.28}$ & 48.57$_{\pm1.13}$/\underline{61.81}$_{\pm6.12}$ & \underline{50.73}$_{\pm1.03}$/56.57$_{\pm3.99}$ & 46.71$_{\pm1.17}$/57.99$_{\pm6.99}$ & 50.11$_{\pm3.83}$/52.01$_{\pm9.29}$ & \textbf{86.43}$_{\pm2.61}$/\textbf{94.40}$_{\pm0.18}$ \\
        Car & \underline{54.31}$_{\pm4.72}$/48.94$_{\pm4.33}$ & 47.76$_{\pm2.08}$/48.13$_{\pm5.92}$ & 49.44$_{\pm2.12}$/\underline{56.01}$_{\pm7.33}$ & 48.20$_{\pm1.80}$/51.26$_{\pm3.79}$ & \textbf{57.16}$_{\pm4.83}$/\textbf{98.99}$_{\pm0.11}$ & \underline{56.07}$_{\pm5.84}$/\underline{51.19}$_{\pm5.65}$ & 47.65$_{\pm0.71}$/45.75$_{\pm4.96}$ & 52.10$_{\pm4.78}$/47.57$_{\pm5.44}$ & 48.71$_{\pm2.76}$/49.84$_{\pm4.05}$ & \textbf{59.87}$_{\pm2.44}$/\textbf{99.02}$_{\pm0.11}$ \\
        Chicken & 47.35$_{\pm1.97}$/\underline{53.31}$_{\pm6.14}$ & \underline{49.25}$_{\pm1.23}$/48.97$_{\pm3.05}$ & 48.93$_{\pm1.87}$/51.30$_{\pm5.26}$ & 47.81$_{\pm3.12}$/47.36$_{\pm4.93}$ & \textbf{79.76}$_{\pm6.15}$/\textbf{84.39}$_{\pm0.92}$ & \underline{51.54}$_{\pm3.01}$/49.61$_{\pm5.84}$ & 49.23$_{\pm1.54}$/51.39$_{\pm3.59}$ & 46.43$_{\pm1.55}$/\underline{54.12}$_{\pm4.00}$ & 47.50$_{\pm2.48}$/52.63$_{\pm5.61}$ & \textbf{77.84}$_{\pm5.55}$/\textbf{84.28}$_{\pm2.07}$ \\
        Diamond & 48.42$_{\pm1.17}$/62.29$_{\pm5.48}$ & \underline{49.19}$_{\pm1.95}$/53.89$_{\pm8.68}$ & 48.25$_{\pm1.11}$/\underline{65.00}$_{\pm3.17}$ & 46.70$_{\pm2.91}$/60.14$_{\pm5.36}$ & \textbf{99.59}$_{\pm0.38}$/\textbf{99.33}$_{\pm0.05}$ & \underline{48.38}$_{\pm1.67}$/56.17$_{\pm3.94}$ & 47.90$_{\pm1.15}$/55.99$_{\pm4.28}$ & 47.09$_{\pm2.41}$/\underline{63.41}$_{\pm5.98}$ & 46.34$_{\pm2.26}$/61.01$_{\pm4.39}$ & \textbf{99.63}$_{\pm0.31}$/\textbf{99.36}$_{\pm0.09}$ \\
        Duck & \underline{50.37}$_{\pm2.80}$/\underline{60.37}$_{\pm7.06}$ & 49.02$_{\pm1.23}$/50.73$_{\pm8.96}$ & 49.73$_{\pm3.16}$/57.87$_{\pm7.48}$ & 48.20$_{\pm3.34}$/57.06$_{\pm5.84}$ & \textbf{83.54}$_{\pm1.83}$/\textbf{97.26}$_{\pm0.33}$ & 48.83$_{\pm1.30}$/61.32$_{\pm8.32}$ & \underline{50.00}$_{\pm2.07}$/47.16$_{\pm3.34}$ & 47.19$_{\pm1.50}$/\underline{64.51}$_{\pm5.61}$ & 49.01$_{\pm2.20}$/48.50$_{\pm3.50}$ & \textbf{83.80}$_{\pm1.43}$/\textbf{97.69}$_{\pm0.31}$ \\
        Fish & 51.16$_{\pm2.04}$/\underline{52.39}$_{\pm6.97}$ & 49.89$_{\pm1.50}$/48.82$_{\pm4.06}$ & \underline{53.09}$_{\pm4.79}$/47.71$_{\pm4.10}$ & 50.59$_{\pm4.65}$/52.14$_{\pm9.45}$ & \textbf{86.15}$_{\pm4.15}$/\textbf{95.93}$_{\pm0.37}$ & \underline{52.97}$_{\pm4.45}$/46.80$_{\pm2.95}$ & 49.80$_{\pm1.63}$/50.90$_{\pm3.64}$ & 51.67$_{\pm4.48}$/44.24$_{\pm1.31}$ & 50.79$_{\pm3.99}$/\underline{51.10}$_{\pm4.32}$ & \textbf{83.94}$_{\pm7.58}$/\textbf{95.91}$_{\pm1.09}$ \\
        Gemstone & 48.56$_{\pm2.51}$/55.85$_{\pm3.84}$ & \underline{52.57}$_{\pm3.69}$/\underline{56.77}$_{\pm3.54}$ & 50.28$_{\pm2.95}$/53.02$_{\pm3.09}$ & 50.16$_{\pm3.50}$/52.26$_{\pm3.16}$ & \textbf{60.68}$_{\pm2.89}$/\textbf{97.84}$_{\pm0.16}$ & 48.71$_{\pm2.36}$/57.48$_{\pm5.93}$ & 48.67$_{\pm1.86}$/56.00$_{\pm5.22}$ & 48.08$_{\pm4.13}$/54.84$_{\pm4.27}$ & \underline{50.55}$_{\pm2.59}$/\underline{57.74}$_{\pm2.92}$ & \textbf{63.69}$_{\pm1.30}$/\textbf{98.04}$_{\pm0.09}$ \\
        Seahorse & 50.49$_{\pm3.37}$/53.10$_{\pm2.72}$ & 48.92$_{\pm0.90}$/48.89$_{\pm7.20}$ & 48.98$_{\pm2.72}$/48.56$_{\pm3.70}$ & \underline{50.90}$_{\pm2.19}$/\underline{55.89}$_{\pm6.50}$ & \textbf{76.88}$_{\pm8.81}$/\textbf{90.81}$_{\pm0.99}$ & 49.60$_{\pm2.29}$/\underline{58.97}$_{\pm3.01}$ & \underline{50.29}$_{\pm1.70}$/52.19$_{\pm4.92}$ & 48.41$_{\pm0.52}$/50.02$_{\pm4.47}$ & 50.20$_{\pm2.05}$/57.46$_{\pm5.51}$ & \textbf{81.56}$_{\pm6.66}$/\textbf{92.79}$_{\pm1.05}$ \\
        Shell & 49.49$_{\pm2.78}$/48.79$_{\pm6.12}$ & \underline{50.04}$_{\pm2.46}$/51.80$_{\pm5.30}$ & 47.07$_{\pm0.71}$/\underline{54.17}$_{\pm7.51}$ & 46.81$_{\pm1.26}$/50.01$_{\pm4.21}$ & \textbf{52.00}$_{\pm8.28}$/\textbf{88.37}$_{\pm0.63}$ & 47.41$_{\pm2.77}$/61.25$_{\pm16.64}$ & \textbf{49.05}$_{\pm1.18}$/54.79$_{\pm9.04}$ & \underline{48.13}$_{\pm2.37}$/\underline{64.17}$_{\pm7.66}$ & 47.12$_{\pm2.67}$/58.06$_{\pm4.53}$ & 44.58$_{\pm7.94}$/\textbf{88.38}$_{\pm1.07}$ \\
        Starfish & 50.71$_{\pm3.08}$/44.65$_{\pm1.56}$ & 51.27$_{\pm3.39}$/49.16$_{\pm7.91}$ & \underline{53.70}$_{\pm2.19}$/\underline{50.21}$_{\pm3.87}$ & 47.65$_{\pm1.91}$/43.92$_{\pm3.35}$ & \textbf{69.82}$_{\pm3.69}$/\textbf{84.80}$_{\pm0.60}$ & \underline{51.87}$_{\pm3.56}$/\underline{55.15}$_{\pm5.38}$ & 48.69$_{\pm1.56}$/37.61$_{\pm4.75}$ & 51.14$_{\pm2.34}$/54.03$_{\pm2.64}$ & 48.40$_{\pm1.52}$/45.28$_{\pm3.10}$ & \textbf{65.42}$_{\pm2.43}$/\textbf{86.56}$_{\pm0.89}$ \\
        Toffees & 47.36$_{\pm2.21}$/51.44$_{\pm2.77}$ & 48.12$_{\pm1.74}$/46.83$_{\pm1.52}$ & \underline{49.62}$_{\pm5.20}$/\underline{54.65}$_{\pm2.57}$ & 45.99$_{\pm3.49}$/50.64$_{\pm2.97}$ & \textbf{67.46}$_{\pm3.71}$/\textbf{95.53}$_{\pm0.79}$ & \underline{48.73}$_{\pm3.98}$/48.40$_{\pm4.24}$ & 46.34$_{\pm0.35}$/49.29$_{\pm3.56}$ & 47.01$_{\pm3.05}$/53.25$_{\pm2.48}$ & 45.12$_{\pm1.87}$/\underline{53.40}$_{\pm3.38}$ & \textbf{65.35}$_{\pm1.95}$/\textbf{96.56}$_{\pm0.56}$ \\
        Average & \underline{49.87}$_{\pm1.84}$/53.81$_{\pm3.57}$ & 49.84$_{\pm1.68}$/49.73$_{\pm3.87}$ & 49.87$_{\pm1.79}$/\underline{55.75}$_{\pm3.67}$ & 48.18$_{\pm1.55}$/51.61$_{\pm3.78}$ & \textbf{72.90}$_{\pm0.85}$/\textbf{92.95}$_{\pm0.22}$ & \underline{50.13}$_{\pm2.39}$/55.48$_{\pm4.08}$ & 48.98$_{\pm1.18}$/48.89$_{\pm5.62}$ & 48.68$_{\pm1.96}$/\underline{55.51}$_{\pm4.22}$ & 48.53$_{\pm1.69}$/52.03$_{\pm5.94}$ & \textbf{72.35}$_{\pm1.12}$/\textbf{93.32}$_{\pm0.11}$ \\
        \hline
    \end{tabular}%
    }
\end{table*}

\begin{table*}[!t]
    \centering
    \caption{Object-AUROC/P-AUROC results under the open-set protocol, where models are trained using two or four known anomaly classes and evaluated on detecting the remaining overall anomaly categories in the Anomaly-ShapeNet dataset. Results are reported as mean $\pm$ standard deviation, with object-level and point-level metrics separated by ``/''.}
    \label{o-ASN}
    \setlength{\tabcolsep}{2pt}
    \renewcommand{\arraystretch}{0.95}
    \resizebox{\textwidth}{!}{%
    \begin{tabular}{c||ccccc|ccccc}
        \hline
        \raisebox{-1.45ex}[0pt][0pt]{Dataset} & & & \multicolumn{1}{c}{Two Sample per class} & & & & & \multicolumn{1}{c}{Four Sample per class} & & \\
        \cline{2-11}
        & DevNet$_{\mathrm{pm}}$ & DevNet$_{\mathrm{pb}}$ & DRA$_{\mathrm{pm}}$ & DRA$_{\mathrm{pb}}$ & Ours & DevNet$_{\mathrm{pm}}$ & DevNet$_{\mathrm{pb}}$ & DRA$_{\mathrm{pm}}$ & DRA$_{\mathrm{pb}}$ & Ours \\
        \hline
        Ashtray & 67.81$_{\pm10.95}$/49.24$_{\pm5.82}$ & 56.86$_{\pm19.78}$/\underline{52.03}$_{\pm3.95}$ & \underline{86.92}$_{\pm6.66}$/47.83$_{\pm4.42}$ & 79.44$_{\pm3.48}$/49.00$_{\pm2.81}$ & \textbf{99.89}$_{\pm0.22}$/\textbf{95.62}$_{\pm0.63}$ & 57.52$_{\pm9.51}$/\underline{52.18}$_{\pm1.54}$ & 50.76$_{\pm11.92}$/49.41$_{\pm2.47}$ & \underline{65.22}$_{\pm7.39}$/50.54$_{\pm2.24}$ & 62.61$_{\pm8.30}$/48.64$_{\pm1.80}$ & \textbf{99.73}$_{\pm0.53}$/\textbf{96.14}$_{\pm1.23}$ \\
        Bag & 59.05$_{\pm5.01}$/\underline{53.83}$_{\pm9.78}$ & 50.48$_{\pm15.87}$/53.37$_{\pm12.35}$ & \underline{66.77}$_{\pm11.32}$/50.05$_{\pm7.34}$ & 66.31$_{\pm7.39}$/46.69$_{\pm4.20}$ & \textbf{87.56}$_{\pm5.94}$/\textbf{96.05}$_{\pm2.17}$ & 52.19$_{\pm5.53}$/47.63$_{\pm2.06}$ & 46.86$_{\pm10.41}$/46.23$_{\pm0.00}$ & \underline{67.89}$_{\pm6.40}$/\underline{53.31}$_{\pm3.73}$ & 55.67$_{\pm6.74}$/50.70$_{\pm4.30}$ & \textbf{86.00}$_{\pm4.52}$/\textbf{96.10}$_{\pm2.18}$ \\
        Bottle & 72.19$_{\pm10.02}$/\underline{55.63}$_{\pm6.61}$ & 63.06$_{\pm12.74}$/53.23$_{\pm5.22}$ & 78.61$_{\pm6.97}$/53.53$_{\pm4.87}$ & \underline{79.30}$_{\pm6.59}$/51.34$_{\pm5.92}$ & \textbf{91.20}$_{\pm1.40}$/\textbf{82.30}$_{\pm0.40}$ & 58.65$_{\pm15.05}$/\underline{52.54}$_{\pm3.48}$ & 45.15$_{\pm11.33}$/50.55$_{\pm2.99}$ & 67.20$_{\pm9.77}$/51.12$_{\pm3.47}$ & \underline{71.56}$_{\pm8.43}$/49.67$_{\pm2.92}$ & \textbf{94.91}$_{\pm1.45}$/\textbf{82.40}$_{\pm0.82}$ \\
        Bowl & 56.97$_{\pm9.60}$/\underline{51.82}$_{\pm5.32}$ & 52.57$_{\pm7.36}$/50.85$_{\pm5.27}$ & \underline{66.65}$_{\pm7.41}$/49.40$_{\pm4.19}$ & 61.06$_{\pm11.03}$/50.59$_{\pm5.38}$ & \textbf{96.66}$_{\pm0.25}$/\textbf{98.81}$_{\pm0.09}$ & 51.64$_{\pm7.33}$/\underline{51.01}$_{\pm4.15}$ & 45.43$_{\pm6.45}$/49.82$_{\pm4.11}$ & \underline{61.29}$_{\pm5.35}$/50.73$_{\pm4.25}$ & 55.56$_{\pm6.64}$/50.12$_{\pm4.40}$ & \textbf{96.84}$_{\pm0.45}$/\textbf{98.86}$_{\pm0.07}$ \\
        Bucket & 59.60$_{\pm7.00}$/\underline{59.10}$_{\pm8.10}$ & 54.38$_{\pm8.41}$/53.61$_{\pm4.62}$ & \underline{71.40}$_{\pm5.39}$/51.17$_{\pm4.60}$ & 67.70$_{\pm6.64}$/49.20$_{\pm3.36}$ & \textbf{87.69}$_{\pm0.65}$/\textbf{84.66}$_{\pm0.77}$ & 58.50$_{\pm9.92}$/50.85$_{\pm1.51}$ & 41.62$_{\pm4.52}$/48.14$_{\pm2.40}$ & \underline{64.47}$_{\pm5.74}$/\underline{51.42}$_{\pm1.83}$ & 59.47$_{\pm10.55}$/50.84$_{\pm2.47}$ & \textbf{87.41}$_{\pm0.82}$/\textbf{84.59}$_{\pm0.38}$ \\
        Cabinet & 61.13$_{\pm8.67}$/47.37$_{\pm1.98}$ & 48.87$_{\pm7.73}$/48.45$_{\pm3.43}$ & \underline{75.86}$_{\pm7.23}$/47.79$_{\pm5.05}$ & 71.33$_{\pm10.17}$/\underline{48.87}$_{\pm2.07}$ & \textbf{84.67}$_{\pm2.91}$/\textbf{94.19}$_{\pm0.16}$ & 54.97$_{\pm8.89}$/\underline{51.97}$_{\pm2.79}$ & 46.25$_{\pm3.84}$/48.79$_{\pm2.16}$ & \underline{75.18}$_{\pm7.75}$/48.93$_{\pm2.93}$ & 68.11$_{\pm4.08}$/48.80$_{\pm1.00}$ & \textbf{81.25}$_{\pm2.92}$/\textbf{93.12}$_{\pm0.29}$ \\
        Cap & 61.22$_{\pm7.74}$/51.41$_{\pm6.83}$ & 55.88$_{\pm7.17}$/51.72$_{\pm5.98}$ & 63.67$_{\pm5.87}$/52.32$_{\pm5.87}$ & \underline{66.84}$_{\pm8.14}$/\underline{53.37}$_{\pm6.08}$ & \textbf{84.12}$_{\pm2.03}$/\textbf{96.38}$_{\pm0.24}$ & 57.70$_{\pm7.73}$/50.64$_{\pm5.07}$ & 55.52$_{\pm7.26}$/50.80$_{\pm5.12}$ & 60.21$_{\pm5.25}$/\underline{50.82}$_{\pm4.14}$ & \underline{61.64}$_{\pm5.66}$/49.92$_{\pm4.79}$ & \textbf{81.27}$_{\pm1.73}$/\textbf{95.37}$_{\pm0.19}$ \\
        Chair & 60.67$_{\pm7.79}$/46.60$_{\pm2.05}$ & 51.13$_{\pm8.50}$/46.59$_{\pm1.78}$ & \textbf{66.14}$_{\pm7.66}$/\underline{47.51}$_{\pm4.91}$ & 57.47$_{\pm6.29}$/47.11$_{\pm3.66}$ & \underline{64.29}$_{\pm3.60}$/\textbf{86.30}$_{\pm0.83}$ & 61.59$_{\pm4.77}$/\underline{50.73}$_{\pm3.22}$ & 54.51$_{\pm5.16}$/50.43$_{\pm3.72}$ & \textbf{68.59}$_{\pm8.92}$/49.26$_{\pm2.52}$ & 61.52$_{\pm9.35}$/49.70$_{\pm2.40}$ & \underline{65.25}$_{\pm3.53}$/\textbf{85.47}$_{\pm0.70}$ \\
        Cup & 54.98$_{\pm4.17}$/\underline{53.14}$_{\pm3.42}$ & 55.24$_{\pm8.46}$/51.79$_{\pm4.10}$ & \underline{63.55}$_{\pm4.20}$/50.37$_{\pm3.43}$ & 59.52$_{\pm7.20}$/48.96$_{\pm4.00}$ & \textbf{99.59}$_{\pm0.18}$/\textbf{97.86}$_{\pm0.32}$ & 48.51$_{\pm6.41}$/49.54$_{\pm3.08}$ & 45.36$_{\pm6.95}$/49.72$_{\pm3.80}$ & \underline{60.06}$_{\pm6.05}$/49.84$_{\pm2.26}$ & 51.69$_{\pm6.35}$/\underline{50.33}$_{\pm3.16}$ & \textbf{99.69}$_{\pm0.11}$/\textbf{98.30}$_{\pm0.23}$ \\
        Desk & 71.28$_{\pm4.69}$/\underline{52.15}$_{\pm5.34}$ & 51.23$_{\pm2.52}$/47.71$_{\pm2.72}$ & \underline{79.30}$_{\pm4.74}$/50.33$_{\pm3.14}$ & 67.75$_{\pm10.44}$/47.52$_{\pm5.00}$ & \textbf{88.22}$_{\pm2.00}$/\textbf{96.59}$_{\pm0.20}$ & \underline{70.20}$_{\pm2.53}$/49.92$_{\pm3.72}$ & 49.07$_{\pm2.28}$/48.82$_{\pm2.27}$ & 67.41$_{\pm9.47}$/49.79$_{\pm1.80}$ & 61.08$_{\pm10.44}$/\underline{50.92}$_{\pm3.41}$ & \textbf{83.50}$_{\pm1.93}$/\textbf{95.96}$_{\pm0.28}$ \\
        Eraser & 46.19$_{\pm10.33}$/\underline{59.19}$_{\pm4.29}$ & \underline{63.24}$_{\pm2.47}$/53.07$_{\pm5.18}$ & 54.87$_{\pm6.35}$/48.26$_{\pm2.84}$ & 62.62$_{\pm4.08}$/48.09$_{\pm5.02}$ & \textbf{99.89}$_{\pm0.22}$/\textbf{97.67}$_{\pm0.67}$ & 42.76$_{\pm8.06}$/\underline{52.76}$_{\pm2.83}$ & \underline{59.43}$_{\pm3.08}$/50.05$_{\pm2.28}$ & 49.06$_{\pm4.19}$/52.44$_{\pm2.44}$ & 57.89$_{\pm3.02}$/50.39$_{\pm2.03}$ & \textbf{100.00}$_{\pm0.00}$/\textbf{98.04}$_{\pm0.34}$ \\
        Headset & 54.05$_{\pm8.54}$/49.34$_{\pm4.98}$ & 51.94$_{\pm8.98}$/51.79$_{\pm3.97}$ & 66.38$_{\pm7.37}$/51.66$_{\pm4.07}$ & \underline{66.95}$_{\pm5.68}$/\underline{52.15}$_{\pm4.14}$ & \textbf{96.17}$_{\pm1.45}$/\textbf{96.00}$_{\pm0.24}$ & 52.00$_{\pm6.59}$/50.08$_{\pm2.82}$ & 46.76$_{\pm7.53}$/49.64$_{\pm3.38}$ & \underline{58.51}$_{\pm7.77}$/\underline{52.85}$_{\pm3.05}$ & 55.84$_{\pm4.13}$/51.02$_{\pm2.05}$ & \textbf{98.07}$_{\pm0.97}$/\textbf{96.63}$_{\pm0.60}$ \\
        Helmet & 54.57$_{\pm6.08}$/\underline{51.76}$_{\pm4.27}$ & 52.24$_{\pm6.23}$/51.70$_{\pm4.68}$ & \underline{65.53}$_{\pm9.11}$/49.83$_{\pm3.72}$ & 56.02$_{\pm6.78}$/48.88$_{\pm3.37}$ & \textbf{81.54}$_{\pm1.22}$/\textbf{95.10}$_{\pm0.11}$ & 54.91$_{\pm3.58}$/\underline{51.22}$_{\pm3.36}$ & 47.01$_{\pm6.43}$/49.80$_{\pm2.76}$ & \underline{58.58}$_{\pm5.34}$/51.07$_{\pm3.54}$ & 50.14$_{\pm6.45}$/50.41$_{\pm3.03}$ & \textbf{79.54}$_{\pm1.28}$/\textbf{95.07}$_{\pm0.19}$ \\
        Jar & 60.48$_{\pm12.08}$/\underline{58.08}$_{\pm2.47}$ & 61.91$_{\pm14.96}$/53.63$_{\pm3.32}$ & \underline{75.90}$_{\pm6.07}$/53.42$_{\pm2.71}$ & 67.44$_{\pm8.22}$/52.06$_{\pm2.07}$ & \textbf{89.00}$_{\pm6.22}$/\textbf{98.12}$_{\pm0.84}$ & 52.57$_{\pm6.09}$/54.62$_{\pm1.27}$ & 47.33$_{\pm9.97}$/\underline{54.93}$_{\pm2.06}$ & 59.83$_{\pm3.27}$/51.49$_{\pm2.01}$ & \underline{62.61}$_{\pm6.92}$/52.00$_{\pm2.22}$ & \textbf{92.00}$_{\pm6.91}$/\textbf{98.54}$_{\pm0.80}$ \\
        Knife & 54.48$_{\pm3.86}$/49.67$_{\pm3.06}$ & 50.59$_{\pm7.48}$/\underline{49.74}$_{\pm2.81}$ & \underline{58.63}$_{\pm4.32}$/49.58$_{\pm3.00}$ & 49.83$_{\pm5.75}$/48.46$_{\pm3.41}$ & \textbf{78.22}$_{\pm1.58}$/\textbf{92.69}$_{\pm0.34}$ & \underline{58.12}$_{\pm6.51}$/49.82$_{\pm2.70}$ & 47.79$_{\pm5.17}$/\underline{49.97}$_{\pm2.14}$ & 55.02$_{\pm5.01}$/48.24$_{\pm2.41}$ & 51.80$_{\pm5.92}$/49.58$_{\pm3.29}$ & \textbf{76.71}$_{\pm2.30}$/\textbf{92.11}$_{\pm0.41}$ \\
        Microphone & 53.09$_{\pm9.02}$/53.87$_{\pm7.11}$ & 47.40$_{\pm11.63}$/\underline{54.00}$_{\pm5.02}$ & \underline{60.74}$_{\pm8.28}$/49.47$_{\pm5.78}$ & 55.12$_{\pm9.75}$/49.51$_{\pm4.65}$ & \textbf{100.00}$_{\pm0.00}$/\textbf{98.36}$_{\pm0.56}$ & 51.65$_{\pm11.45}$/51.42$_{\pm4.33}$ & 54.75$_{\pm10.27}$/\underline{53.11}$_{\pm4.53}$ & \underline{58.41}$_{\pm9.16}$/50.91$_{\pm4.17}$ & 54.56$_{\pm8.09}$/52.91$_{\pm4.42}$ & \textbf{100.00}$_{\pm0.00}$/\textbf{98.59}$_{\pm0.57}$ \\
        Screen & 61.40$_{\pm9.64}$/\underline{52.39}$_{\pm6.37}$ & 45.21$_{\pm8.96}$/48.66$_{\pm4.35}$ & \underline{66.35}$_{\pm4.91}$/48.18$_{\pm5.14}$ & 54.39$_{\pm7.89}$/46.71$_{\pm5.47}$ & \textbf{69.84}$_{\pm4.02}$/\textbf{94.59}$_{\pm0.50}$ & 58.00$_{\pm11.72}$/50.39$_{\pm1.43}$ & 53.94$_{\pm6.26}$/49.52$_{\pm3.68}$ & \underline{70.79}$_{\pm7.49}$/50.34$_{\pm4.36}$ & 62.04$_{\pm8.51}$/\underline{52.40}$_{\pm3.45}$ & \textbf{71.43}$_{\pm6.89}$/\textbf{94.30}$_{\pm0.66}$ \\
        Shelf & 62.29$_{\pm2.95}$/48.16$_{\pm1.74}$ & 60.75$_{\pm3.06}$/47.11$_{\pm2.63}$ & 63.40$_{\pm2.47}$/49.61$_{\pm2.41}$ & \underline{65.76}$_{\pm2.17}$/\underline{49.87}$_{\pm1.92}$ & \textbf{76.44}$_{\pm0.68}$/\textbf{91.84}$_{\pm1.03}$ & 60.08$_{\pm5.50}$/50.32$_{\pm4.03}$ & 58.17$_{\pm2.23}$/49.25$_{\pm2.15}$ & 62.54$_{\pm2.85}$/50.93$_{\pm3.75}$ & \underline{62.70}$_{\pm2.41}$/\underline{52.54}$_{\pm4.50}$ & \textbf{73.54}$_{\pm0.57}$/\textbf{91.94}$_{\pm1.03}$ \\
        Tap & 52.16$_{\pm6.09}$/\underline{51.25}$_{\pm3.00}$ & 49.05$_{\pm6.87}$/49.90$_{\pm3.15}$ & \underline{66.29}$_{\pm8.50}$/49.26$_{\pm4.60}$ & 50.24$_{\pm6.30}$/49.21$_{\pm4.59}$ & \textbf{69.84}$_{\pm2.60}$/\textbf{88.74}$_{\pm1.36}$ & 56.94$_{\pm7.20}$/\underline{50.93}$_{\pm4.89}$ & 51.53$_{\pm6.63}$/49.68$_{\pm4.18}$ & \underline{61.03}$_{\pm6.98}$/50.01$_{\pm2.81}$ & 47.12$_{\pm6.09}$/49.44$_{\pm2.73}$ & \textbf{69.70}$_{\pm3.78}$/\textbf{89.46}$_{\pm1.46}$ \\
        Vase & 57.79$_{\pm7.20}$/\underline{53.06}$_{\pm5.66}$ & 56.45$_{\pm7.47}$/50.98$_{\pm4.72}$ & \underline{67.65}$_{\pm7.45}$/50.68$_{\pm5.46}$ & 54.56$_{\pm7.23}$/50.21$_{\pm5.03}$ & \textbf{88.68}$_{\pm0.39}$/\textbf{96.02}$_{\pm0.27}$ & 53.16$_{\pm7.26}$/\underline{51.64}$_{\pm5.78}$ & 53.17$_{\pm7.23}$/50.04$_{\pm4.55}$ & \underline{59.81}$_{\pm6.62}$/50.67$_{\pm3.52}$ & 53.46$_{\pm5.26}$/51.22$_{\pm4.16}$ & \textbf{87.78}$_{\pm0.43}$/\textbf{95.59}$_{\pm0.44}$ \\
        Average & 59.07$_{\pm7.57}$/\underline{52.35}$_{\pm3.50}$ & 53.92$_{\pm8.83}$/51.00$_{\pm2.25}$ & \underline{68.23}$_{\pm6.61}$/50.01$_{\pm1.70}$ & 62.98$_{\pm7.06}$/49.39$_{\pm1.78}$ & \textbf{86.67}$_{\pm0.78}$/\textbf{93.89}$_{\pm0.09}$ & 55.58$_{\pm7.58}$/\underline{51.01}$_{\pm1.40}$ & 50.02$_{\pm6.75}$/49.93$_{\pm1.69}$ & \underline{62.55}$_{\pm6.54}$/50.74$_{\pm1.21}$ & 58.35$_{\pm6.67}$/50.58$_{\pm1.16}$ & \textbf{86.23}$_{\pm0.84}$/\textbf{93.83}$_{\pm0.09}$ \\
        \hline
    \end{tabular}%
    }
\end{table*}

\section{Experiments}
\subsection{Experiment setup}
\subsubsection{Datasets.} Our benchmark comprises the proposed Open-Industry dataset, as well as the widely used Real3D-AD~\cite{liu2023real3d} and Anomaly-ShapeNet~\cite{Li_2024_CVPR} datasets. (1) Real3D-AD is a high-precision, real-world 3D anomaly detection dataset covering 12 categories, where each category contains only 4 training samples and 100 test samples; the training data are complete 360° surface point clouds reconstructed via multi-view acquisition, stitching, and manual calibration, while the test data are single-view point clouds with 2–3 anomaly types per category, leading to a pronounced domain gap that makes anomaly detection and localization more challenging. (2) Anomaly-ShapeNet is built from synthetic ShapeNetCore2~\cite{shapenet} samples. We use the full set of 50 diverse categories, comprising over 2,000 complete surface point clouds. Each category provides only four normal training samples, while the test set includes both normal and anomalous samples with 2–4 anomaly types, posing a challenging evaluation of robustness and generalization. 
To accommodate our open-set anomaly detection setting, we have detailed the data processing procedures in the \textit{Supplementary Materials}.

\subsubsection{Baseline Description}
In open-set image anomaly detection, images are first converted into feature tokens, and subsequent modeling is performed at the token level. Following this paradigm, we construct 3D baselines by replacing the image encoder with a point cloud backbone and introducing the necessary geometric adaptations. Following M3DM and Reg3D-AD, we adopt a point transformer~\cite{zhao2021point} initialized with PointMAE~\cite{pang2022masked} or PointBERT~\cite{yu2021pointbert} weights. We adapt two representative open-set methods, DevNet~\cite{pang2019deep} and DRA~\cite{ding2022catching}, yielding DevNet${\text{pm}}$, DevNet${\text{pb}}$, DRA${\text{pm}}$, and DRA${\text{pb}}$. DevNet learns anomaly scores through deviation learning with a few labeled anomalies, whereas DRA disentangles abnormal representations to generalize to unseen anomaly types. These baselines allow us to assess the transferability of classical open-set anomaly detection methods to realistic 3D industrial point clouds. Implementation details and hyperparameters are provided in the \textit{Supplementary Materials}.

\subsubsection{Implementation details.}
Following prior works, we optimize the network using the AdamW optimizer~\cite{loshchilov2017decoupled} with an initial learning rate of $10^{-3}$, and adopt a cosine learning rate schedule~\cite{li2019exponential} that keeps the learning rate constant for the first 50 epochs and then gradually decays it to $0.1$ over the next 400 training epochs. By default, the multi-scale ranges are set to 40, 80, and 120, and the number of aggregation points is fixed at 128 following Simple3D~\cite{cheng2025towards}. Regarding the hyperparameters, we set the target cardinality $\mathcal{N}=1000$ to balance memory efficiency and detection performance. For computational simplicity, we set $\mathcal{K}=3$ for nearest neighbor search. The fusion coefficient is empirically fixed at $\gamma=0.3$ to ensure a balanced contribution from dual-distribution manifolds during inference. 
All comparison experiments, ablation experiments and sensitivity analyses were evaluated using five-fold cross-validation, and the mean $\pm$ standard deviation across 80 categories is reported.

\subsection{Main Results}

\subsubsection{Results on Open-Industry}
Table~\ref{o-open} reports the object-level and point-level results on Open-Industry. Under the 5-samples-per-class setting, our method achieves the best average performance with 84.39\% O-AUROC and 70.52\% P-AUROC, surpassing the second-best baseline by 17.31\% and 16.17\%, respectively. Under the 10-samples-per-class setting, our method again ranks first with 82.74\% O-AUROC and 76.29\% P-AUROC, outperforming the second-best method by 11.96\% and 20.55\%, respectively. 

\subsubsection{Results on Real3D-AD}
The object-level results on Real3D-AD are shown in Table~\ref{o-real3dad}. Our method achieves the best average O-AUROC of 72.90\% and 72.35\% under the 2-samples-per-class and 4-samples-per-class settings, respectively, while the adapted baselines remain close to chance level on average. This indicates that our method is more robust under severe train--test discrepancy and extremely limited supervision.

\subsubsection{Results on Anomaly-ShapeNet}
Table~\ref{o-ASN} reports the object-level results on Anomaly-ShapeNet. Our method consistently achieves the best average performance, reaching 86.67\% O-AUROC under the 2-samples-per-class setting and 86.23\% under the 4-samples-per-class setting. The results further verify the generalization ability in open-set supervised 3D anomaly detection.

\subsection{Subsampling Method Analysis}
The proposed Correspondence Distributions Subsampling is designed to eliminate ambiguous features in the overlapping regions between the normal and abnormal distributions, thereby enhancing discriminability. In Table~\ref{tab:subsampling_open_industry}, we compare the proposed method and its variants on Open-Industry. Specifically, $M_1$ denotes Correspondence Distributions Subsampling, $M_2$ denotes Identity Sampling, $M_3$ denotes Random Sampling, $M_4$ denotes Greedy Coreset Sampling, and $M_5$ denotes Approximate Greedy Coreset Sampling. As shown in Table~\ref{tab:subsampling_open_industry}, $M_1$ achieves the best performance, with 84.39 O-AUROC and 70.52 P-AUROC, and it brings gains of 4.29 in O-AUROC and 3.09 in P-AUROC compared with Identity Sampling. This suggests that our method better removes ambiguous features in the overlap between normal and abnormal distributions, leading to a clearer separation boundary and stronger discriminability.

\begin{table}[!ht]
\centering
\caption{Performance comparison of different subsampling strategies on Open-Industry under 5 samples supervised.}
\label{tab:subsampling_open_industry}
\resizebox{0.8\columnwidth}{!}{%
\begin{tabular}{c||cccc}
\hline
Method & Time Complexity & O-AUROC & P-AUROC & FPS\\
\hline
$M_1$ & $\mathcal{O}(mN)$ & \textbf{84.39{\scriptsize $\pm$0.44}}& \textbf{70.52{\scriptsize $\pm$0.56}}& 8.4\\
$M_2$ & $\mathcal{O}(1)$ & 80.10{\scriptsize $\pm$1.32} & 67.43{\scriptsize $\pm$2.40}  & \textbf{10.1}\\
$M_3$ & $\mathcal{O}(N)$ & 81.33{\scriptsize $\pm$1.88} & 68.98{\scriptsize $\pm$2.18} & 9.6\\
$M_4$ & $\mathcal{O}(N^2)$ & 83.20{\scriptsize $\pm$0.95} & 69.84{\scriptsize $\pm$2.21} & 8.8\\
$M_5$ & $\mathcal{O}(mN)$ & 83.50{\scriptsize $\pm$4.69} & 69.66{\scriptsize $\pm$6.35} & 9.0\\
\hline
\end{tabular}%
}

\end{table}

\subsection{Ablation Study}
We conduct an ablation study on Open-Industry under the 5-sample-per-class setting. Here, $M_6$ is the normal-only baseline; $M_7$ adds the anomalous memory bank and expanded normal memory; $M_8$ further incorporates simulated anomalies; and $M_9$ denotes the full model with correspondence distribution subsampling. As shown in Table~\ref{tab:ablation_open_industry}, performance improves progressively across all variants. Explicit anomalous modeling enhances both object- and point-level detection, simulated anomalies further enrich the abnormal feature space, and support refinement yields the best overall results. These findings confirm the effectiveness of dual-distribution modeling, anomaly augmentation, and correspondence-based subsampling. Complete parameter sensitivity results are provided in the \textit{Supplementary Materials}.

\begin{table}[!ht]
\centering
\caption{Ablation study on Open-Industry under the 5 samples per-class supervision setting.}
\label{tab:ablation_open_industry}
\resizebox{0.6\columnwidth}{!}{%
\begin{tabular}{c||cc}
\hline
Method & O-AUROC& P-AUROC\\
\hline
$M_6$ & 83.08{\scriptsize $\pm$0.59}
& 69.18{\scriptsize $\pm$0.40} \\

$M_7$ & 83.16{\scriptsize $\pm$0.29}
& 70.17{\scriptsize $\pm$0.79} \\

$M_8$ & 83.21{\scriptsize $\pm$0.58}
& 70.25{\scriptsize $\pm$0.24} \\

$M_9$ & \textbf{84.39}{\scriptsize $\pm$0.44}
& \textbf{70.52}{\scriptsize $\pm$0.56} \\
\hline
\end{tabular}%
}
\end{table}

\subsection{Feature Distribution Analysis}
\begin{figure}[!ht]
    \centering   \includegraphics[width=0.7\linewidth]{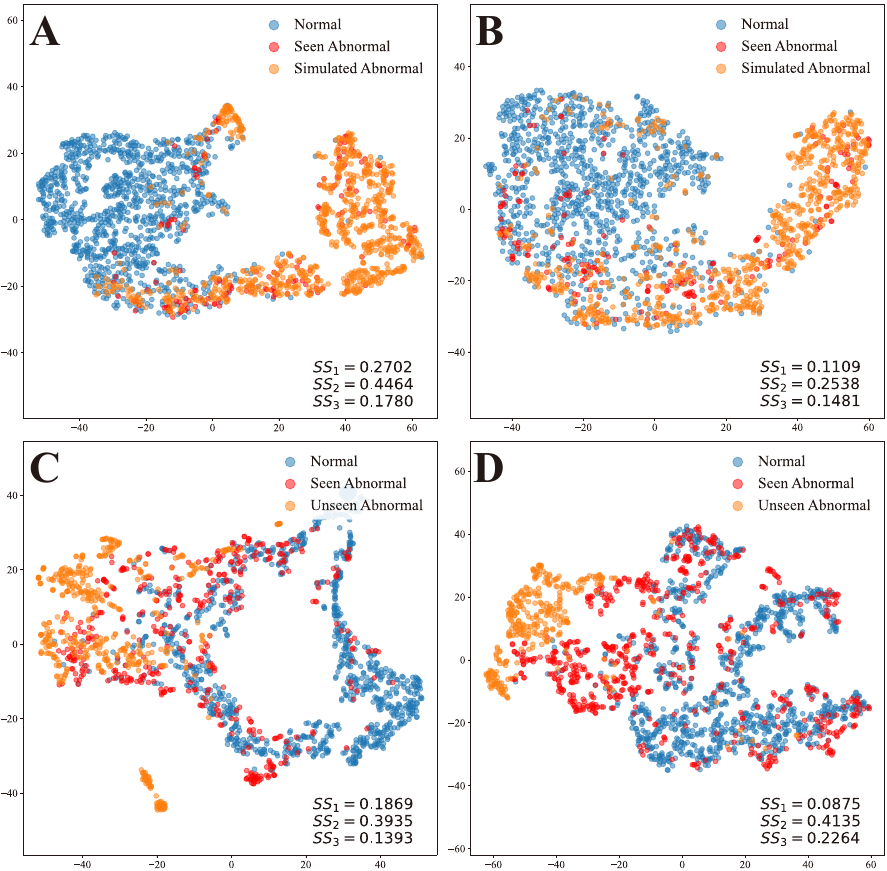}
    \caption{Feature distribution visualizations in the training and inference stages. A and B show selected training features of Normal, Seen Abnormal, and Simulated Abnormal. C and D show selected test features of Normal, Seen Abnormal, and Unseen Abnormal. SS denotes the Silhouette Score. In A and B, $SS_1$, $SS_2$, and $SS_3$ correspond to Normal vs. Seen Abnormal, Normal vs. Simulated Abnormal, and Seen Abnormal vs. Simulated Abnormal, respectively. In C and D, they correspond to Normal vs. Seen Abnormal, Normal vs. Unseen Abnormal, and Seen Abnormal vs. Unseen Abnormal.}
    \label{Distribution}
\end{figure}

Figure~\ref{Distribution} (A, B) shows that simulated anomalies effectively expand the abnormal feature space while remaining largely separated from normal features, indicating a well-decoupled normal--abnormal representation. Limited overlap arises from geometrically similar parts with different semantic priors, which cannot be fully distinguished by geometry-only operators and remains a direction for future work.
Figure~\ref{Distribution} (C, D) further reveals a clear distribution shift between seen and unseen anomalies, as supported by the relatively high Silhouette Scores. Despite this shift, the learned representations remain discriminative for unseen defects, demonstrating open-set generalization beyond the observed anomaly categories. FPS measures the end-to-end processing speed after data loading.


\section{Conclusion}
We introduce open-set supervised 3D anomaly detection, where models are trained with normal samples and a small number of known anomalies while detecting unseen anomaly categories at test time. To support this setting, we present Open-Industry, a real-world industrial point cloud dataset, and establish a benchmark for open-set supervised 3D anomaly detection. We further propose Open3D-AD, a dual-distribution framework that models normal and anomalous feature manifolds to improve generalization to unseen anomalies. Experiments on Open-Industry, Real3D-AD, and Anomaly-ShapeNet demonstrate the effectiveness of our method.


\bibliography{references}
\end{document}